\documentclass[nohyperref]{article}

\usepackage{microtype}
\usepackage{graphicx}
\usepackage{subfigure}
\usepackage{booktabs} 

\usepackage{hyperref}

\usepackage[accepted]{icml2022}

\usepackage{amsmath}
\usepackage{amssymb}
\usepackage{mathtools}
\usepackage{amsthm}
\usepackage{multirow}
\usepackage{enumitem}
\usepackage{xcolor}     

\usepackage[capitalize,noabbrev]{cleveref}

\theoremstyle{plain}

\theoremstyle{definition}

\theoremstyle{remark}

\usepackage[textsize=tiny]{todonotes}

\usepackage{amsmath,amsfonts,bm}
\usepackage{xparse}

\DeclareDocumentCommand\W{ g g }{%
        \IfNoValueTF {#1} {\mathbf{W}} {
            \IfNoValueTF {#2} {\mathbf{W}^{(#1)}}{\mathbf{W}^{(#1)}_{#2}}
        }
}

\DeclareDocumentCommand\bias{ g g }{%
        \IfNoValueTF {#1} {\mathbf{b}} {
            \IfNoValueTF {#2} {\mathbf{b}^{(#1)}}{\mathbf{b}^{(#1)}_{#2}}
        }
}

\DeclareDocumentCommand\betavar{ g g }{%
        \IfNoValueTF {#1} {\bm{\beta}} {
            \IfNoValueTF {#2} {{\bm{\beta}^{(#1)}}{}}{\bm{\beta}^{(#1)}_{#2}}
        }
}

\DeclareDocumentCommand\xivar{ g g }{%
        \IfNoValueTF {#1} {\bm{\xi}} {
            \IfNoValueTF {#2} {{\bm{\xi}^{(#1)}}{}}{\bm{\xi}^{(#1)}_{#2}}
        }
}

\DeclareDocumentCommand\xivarn{ g g }{%
        \IfNoValueTF {#1} {\bm{\xi^-}} {
            \IfNoValueTF {#2} {\bm{\xi^-}^{+(#1)}}{\bm{\xi^-}^{+(#1)}_{#2}}
        }
}

\DeclareDocumentCommand\xivarp{ g g }{%
        \IfNoValueTF {#1} {\bm{\xi^+}} {
            \IfNoValueTF {#2} {\bm{\xi^+}^{+(#1)}}{\bm{\xi^+}^{+(#1)}_{#2}}
        }
}

\DeclareDocumentCommand\nuvar{ g g }{%
        \IfNoValueTF {#1} {{\bm{\nu}}} {
            \IfNoValueTF {#2} {{\bm{\nu}^{(#1)}}{}}{\nu^{(#1)}_{#2}{}}
        }
}

\DeclareDocumentCommand\hnuvar{ g g }{%
        \IfNoValueTF {#1} {\bm{\hat{\nu}}} {
            \IfNoValueTF {#2} {{\bm{\hat{\nu}}^{(#1)}}{}}{\hat{\nu}^{(#1)}_{#2}{}}
        }
}

\DeclareDocumentCommand\muvar{ g g }{%
        \IfNoValueTF {#1} {\bm{\mu}} {
            \IfNoValueTF {#2} {{\bm{\mu}^{(#1)}}{}}{\mu^{(#1)}_{#2}}
        }
}

\DeclareDocumentCommand\tauvar{ g g }{%
        \IfNoValueTF {#1} {\bm{\tau}} {
            \IfNoValueTF {#2} {{\bm{\tau}^{(#1)}}{}}{\tau^{(#1)}_{#2}}
        }
}

\DeclareDocumentCommand\pivar{ g g }{%
        \IfNoValueTF {#1} {\bm{\pi}} {
            \IfNoValueTF {#2} {{\bm{\pi}^{(#1)}}{}}{\pi^{(#1)}_{#2}}
        }
}

\DeclareDocumentCommand\gammavar{ g g }{%
        \IfNoValueTF {#1} {\bm{\gamma}} {
            \IfNoValueTF {#2} {{\bm{\gamma}^{(#1)}}{}}{\gamma^{(#1)}_{#2}}
        }
}

\DeclareDocumentCommand\lambdavar{ g g }{%
        \IfNoValueTF {#1} {\bm{\lambda}} {
            \IfNoValueTF {#2} {{\bm{\lambda}^{(#1)}}{}}{\lambda^{(#1)}_{#2}}
        }
}

\DeclareDocumentCommand\tbetavar{ g g }{%
        \IfNoValueTF {#1} {{\bm{\tilde{\beta}}}} {
            \IfNoValueTF {#2} {{{\bm{\tilde{\beta}}}^{(#1)}}{}}{{{\tilde{\beta}}^{(#1)}_{#2}}}
        }
}

\DeclareDocumentCommand\alphavar{ g g }{%
        \IfNoValueTF {#1} {\bm{\alpha}} {
            \IfNoValueTF {#2} {{\bm{\alpha}^{(#1)}}}{\alpha^{(#1)}_{#2}}
        }
}

\DeclareDocumentCommand\zcut{ g g }{%
        \IfNoValueTF {#1} {\bm{q}} {
            \IfNoValueTF {#2} {{\bm{q}^{(#1)}}}{q^{(#1)}_{#2}}
        }
}

\DeclareDocumentCommand\Zcut{ g g }{%
        \IfNoValueTF {#1} {\bm{Q}} {
            \IfNoValueTF {#2} {{\bm{Q}^{(#1)}}}{\bm{Q}^{(#1)}_{#2}}
        }
}

\DeclareDocumentCommand\xcut{ g g }{%
        \IfNoValueTF {#1} {\bm{h}} {
            \IfNoValueTF {#2} {{\bm{h}^{(#1)}}}{h^{(#1)}_{#2}}
        }
}

\DeclareDocumentCommand\Xcut{ g g }{%
        \IfNoValueTF {#1} {\bm{H}} {
            \IfNoValueTF {#2} {{\bm{H}^{(#1)}}}{\bm{H}^{(#1)}_{#2}}
        }
}

\DeclareDocumentCommand\hxcut{ g g }{%
        \IfNoValueTF {#1} {\bm{g}} {
            \IfNoValueTF {#2} {{\bm{g}^{(#1)}}}{g^{(#1)}_{#2}}
        }
}

\DeclareDocumentCommand\hXcut{ g g }{%
        \IfNoValueTF {#1} {\bm{G}} {
            \IfNoValueTF {#2} {{\bm{G}^{(#1)}}}{\bm{G}^{(#1)}_{#2}}
        }
}

\DeclareDocumentCommand\D{ g g }{%
        \IfNoValueTF {#1} {\mathbf{D}} {
            \IfNoValueTF {#2} {\mathbf{D}^{(#1)}}{\mathbf{D}^{(#1)}_{#2}}
        }
}

\DeclareDocumentCommand\A{ g g }{%
        \IfNoValueTF {#1} {\mathbf{A}} {
            \IfNoValueTF {#2} {\mathbf{A}^{(#1)}}{\mathbf{A}^{(#1)}_{#2}}
        }
}

\DeclareDocumentCommand\a{ g g }{%
        \IfNoValueTF {#1} {\mathbf{a}} {
            \IfNoValueTF {#2} {\mathbf{a}^{(#1)}}{{a}^{(#1)}_{#2}}
        }
}

\DeclareDocumentCommand\al{ g g }{%
        \IfNoValueTF {#1} {\underline{\mathbf{a}}} {
            \IfNoValueTF {#2} {\underline{\mathbf{a}}^{(#1)}}{\underline{a}^{(#1)}_{#2}}
        }
}

\DeclareDocumentCommand\au{ g g }{%
        \IfNoValueTF {#1} {\overline{\mathbf{a}}} {
            \IfNoValueTF {#2} {\overline{\mathbf{a}}^{(#1)}}{\overline{a}^{(#1)}_{#2}}
        }
}

\DeclareDocumentCommand\c{ g }{%
        \IfNoValueTF {#1} {\bm{c}} {
            {c^{({#1})}}
        }
}

\DeclareDocumentCommand\cl{ g }{%
        \IfNoValueTF {#1} {\underline{c}} {
            {\underline{c}^{({#1})}}
        }
}

\DeclareDocumentCommand\cu{ g }{%
        \IfNoValueTF {#1} {\overline{c}} {
            {\overline{c}^{({#1})}}
        }
}

\DeclareDocumentCommand\AA{ g g }{
        \IfNoValueTF {#1} {\mathbf{\Omega}} {
            \IfNoValueTF {#2} {\mathbf{\Omega}(#1, #1)}{\mathbf{\Omega}(#1, #2)}
        }
}

\DeclareDocumentCommand\S{ g g }{%
        \IfNoValueTF {#1} {\mathbf{S}} {
            \IfNoValueTF {#2} {\mathbf{S}^{(#1)}}{\mathbf{S}^{(#1)}_{#2}}
        }
}

\DeclareDocumentCommand\K{ g g }{%
        \IfNoValueTF {#1} {\mathbf{K}} {
            \IfNoValueTF {#2} {\mathbf{K}^{(#1)}}{\mathbf{K}^{(#1)}_{#2}}
        }
}

\DeclareDocumentCommand\B{ g g }{%
        \IfNoValueTF {#1} {\mathbf{B}} {
            \IfNoValueTF {#2} {\mathbf{B}^{(#1)}}{\mathbf{B}^{(#1)}_{#2}}
        }
}

\DeclareDocumentCommand\lowerb{ g g }{%
        \IfNoValueTF {#1} {{\mathbf{\underline{b}}}} {
            \IfNoValueTF {#2} {{\mathbf{\underline{b}}}^{(#1)}}{{\mathbf{\underline{b}}}^{(#1)}_{#2}}
        }
}

\DeclareDocumentCommand\z{ g g }{%
        \IfNoValueTF {#1} {\mathbf{z}} {
            \IfNoValueTF {#2} {\mathbf{z}^{(#1)}}{z^{(#1)}_{#2}}
        }
}

\DeclareDocumentCommand\hz{ g g }{%
        \IfNoValueTF {#1} {\hat{\mathbf{z}}} {
            \IfNoValueTF {#2} {\hat{\mathbf{z}}^{(#1)}}{\hat{z}^{(#1)}_{#2}}
        }
}

\DeclareDocumentCommand\hx{ g g }{%
        \IfNoValueTF {#1} {\hat{\boldsymbol{x}}} {
            \IfNoValueTF {#2} {\hat{\boldsymbol{x}}^{(#1)}}{\hat{x}^{(#1)}_{#2}}
        }
}

\DeclareDocumentCommand\x{ g g }{%
        \IfNoValueTF {#1} {\boldsymbol{x}} {
            \IfNoValueTF {#2} {\boldsymbol{x}^{(#1)}}{x^{(#1)}_{#2}}
        }
}

\DeclareDocumentCommand\bu{ g g }{%
        \IfNoValueTF {#1} {\boldsymbol{u}} {
            \IfNoValueTF {#2} {\boldsymbol{u}^{(#1)}}{{u}^{(#1)}_{#2}}
        }
}

\DeclareDocumentCommand\buh{ g g }{%
        \IfNoValueTF {#1} {\boldsymbol{u}^*} {
            \IfNoValueTF {#2} {\boldsymbol{{u}}^{*(#1)}}{{{u}}^{*(#1)}_{#2}}
        }
}

\DeclareDocumentCommand\bl{ g g }{%
        \IfNoValueTF {#1} {\boldsymbol{l}} {
            \IfNoValueTF {#2} {\boldsymbol{l}^{(#1)}}{{l}^{(#1)}_{#2}}
        }
}

\DeclareDocumentCommand\blh{ g g }{%
        \IfNoValueTF {#1} {\boldsymbol{l}^*} {
            \IfNoValueTF {#2} {\boldsymbol{l}^{*(#1)}}{{{l}}^{*(#1)}_{#2}}
        }
}

\DeclareDocumentCommand\aaa{ g }{%
        \IfNoValueTF {#1} {\bm{a}} {
            {\bm{a}^{({#1})}}
        }
}

\DeclareDocumentCommand\haaa{ g }{%
        \IfNoValueTF {#1} {\bm{\hat{a}}} {
            {\bm{\hat{a}}^{({#1})}}
        }
}

\DeclareDocumentCommand\bbb{ g g }{%
        \IfNoValueTF {#1} {\mathbf{P}} {
            \IfNoValueTF {#2} {{\mathbf{P}_{#1}}}{{\mathbf{P}_{#1}^{({#2})}}}
        }
}

\DeclareDocumentCommand\hbbb{ g g }{%
        \IfNoValueTF {#1} {\mathbf{\hat{P}}} {
            \IfNoValueTF {#2} {{\mathbf{\hat{P}}_{#1}}}{{\mathbf{\hat{P}}_{#1}^{({#2})}}}
        }
}

\DeclareDocumentCommand\ccc{ g g }{%
        \IfNoValueTF {#1} {\mathbf{q}} {
            \IfNoValueTF {#2} {{\mathbf{q}_{#1}}}{{\mathbf{q}_{#1}^{(#2)}}{}}
        }
}

\DeclareDocumentCommand\constc{ g }{%
        \IfNoValueTF {#1} {c} {
            {c^{({#1})}}
        }
}

\DeclareDocumentCommand\setz{ g g }{%
        \IfNoValueTF {#1} {\mathcal{Z}} {
            \IfNoValueTF {#2} {\mathcal{Z}^{(#1)}}{\mathcal{Z}^{(#1)}_{#2}}
        }
}

\DeclareDocumentCommand\setzp{ g g }{%
        \IfNoValueTF {#1} {\mathcal{Z^+}} {
            \IfNoValueTF {#2} {\mathcal{Z}^{+(#1)}}{\mathcal{Z}^{+(#1)}_{#2}}
        }
}

\DeclareDocumentCommand\setzn{ g g }{%
        \IfNoValueTF {#1} {\mathcal{Z^-}} {
            \IfNoValueTF {#2} {\mathcal{Z}^{-(#1)}}{\mathcal{Z}^{-(#1)}_{#2}}
        }
}

\DeclareDocumentCommand\tsetz{ g g }{%
        \IfNoValueTF {#1} {\tilde{\mathcal{Z}}} {
            \IfNoValueTF {#2} {\tilde{\mathcal{Z}}^{(#1)}}{\tilde{\mathcal{Z}}^{(#1)}_{#2}}
        }
}

\DeclareDocumentCommand\seti{ g g }{%
        \IfNoValueTF {#1} {\mathcal{I}} {
            \IfNoValueTF {#2} {\mathcal{I}^{(#1)}}{\mathcal{I}^{(#1)}_{#2}}
        }
}

\DeclareDocumentCommand\setip{ g g }{%
        \IfNoValueTF {#1} {\mathcal{I}^{+}} {
            \IfNoValueTF {#2} {\mathcal{I}^{+(#1)}}{\mathcal{I}^{+(#1)}_{#2}}
        }
}

\DeclareDocumentCommand\setin{ g g }{%
        \IfNoValueTF {#1} {\mathcal{I}^{-}} {
            \IfNoValueTF {#2} {\mathcal{I}^{-(#1)}}{\mathcal{I}^{-(#1)}_{#2}}
        }
}

\DeclareDocumentCommand\tseti{ g g }{%
        \IfNoValueTF {#1} {\tilde{\mathcal{I}}} {
            \IfNoValueTF {#2} {\tilde{\mathcal{I}}^{(#1)}}{\tilde{\mathcal{I}}^{(#1)}_{#2}}
        }
}

\DeclareDocumentCommand\tz{ g g }{%
        \IfNoValueTF {#1} {\tilde{z}} {
            \IfNoValueTF {#2} {\tilde{z}^{(#1)}}{\tilde{z}^{(#1)}_{#2}}
        }
}

\DeclareDocumentCommand\f{ g g }{%
        \IfNoValueTF {#1} {f} {
            \IfNoValueTF {#2} {f^{(#1)}}{f^{(#1)}_{#2}}
        }
}

\DeclareDocumentCommand\lf{ g g }{%
        \IfNoValueTF {#1} {\underline{f}} {
            \IfNoValueTF {#2} {\underline{f}^{(#1)}}{\underline{f}^{(#1)}_{#2}}
        }
}

\def\eqref#1{(\ref{#1})}

\def\1{\bm{1}}

\DeclareMathAlphabet{\mathsfit}{\encodingdefault}{\sfdefault}{m}{sl}
\SetMathAlphabet{\mathsfit}{bold}{\encodingdefault}{\sfdefault}{bx}{n}

\newcommand{\todob}[1]{{\color{blue} #1}}
\newcommand{\todor}[1]{{\color{red} #1}}

\icmltitlerunning{Toward Robust Spiking Neural Network Against Adversarial Perturbation}

\begin{document}

\twocolumn[
\icmltitle{Toward Robust Spiking Neural Network Against Adversarial Perturbation}



\icmlsetsymbol{equal}{*}

\begin{icmlauthorlist}
\icmlauthor{Ling Liang}{ucsb}
\icmlauthor{Kaidi Xu}{drexel}
\icmlauthor{Xing Hu}{ict}
\icmlauthor{Lei Deng}{tsinghua}
\icmlauthor{Yuan Xie}{ucsb}
\end{icmlauthorlist}


\centering
\vskip 0.1in
\textsuperscript{1} University of California, Santa Barbara \\
\textsuperscript{2} Drexel University \\
\textsuperscript{3} Institute of Computing Technology, CAS \\
\textsuperscript{4} Tsinghua University \\

\vskip 0.05in
{\tt\small liangling@ucsb.edu, kx46@drexel.edu, huxing@ict.ac.cn, leideng@mail.tsinghua.edu.cn, yuanxie@ucsb.edu }



\vskip 0.3in
]




\begin{abstract}
As spiking neural networks (SNNs) are deployed increasingly in real-world efficiency critical applications,  the security concerns in SNNs attract more attention.
Currently, researchers have already demonstrated an SNN can be attacked with adversarial examples. How to build a robust SNN becomes an urgent issue.
Recently, many studies apply certified training in artificial neural networks (ANNs), which can improve the robustness of an NN model promisely. However, existing certifications cannot transfer to SNNs directly because of the distinct neuron behavior and input formats for SNNs. In this work, we first design S-IBP and S-CROWN that tackle the non-linear functions in SNNs' neuron modeling. Then, we formalize the boundaries for both digital and spike inputs. Finally, we demonstrate the efficiency of our proposed robust training method in different datasets and model architectures. Based on our experiment, we can achieve a maximum $37.7\%$ attack error reduction with $3.7\%$ original accuracy loss. To the best of our knowledge, this is the first analysis on robust training of SNNs.

\end{abstract}

\section{Introduction}

Spiking neural networks (SNNs) are a class of models that have the potential ability to simulate the behavior of neuron circuits in the brain. Except for the biological character of SNN models, the event-driven based information propagation mechanism makes SNN can be deployed with limited computation resources. Many neuromorphic chips are designed to achieve low power SNN inference, such as TrueNorth \cite{akopyan2015truenorth}, Loihi\cite{davies2018loihi}, SpinalFlow~\cite{narayanan2020spinalflow} and H2Learn~\cite{liang2021h2learn}. 

With more attention to the study of SNNs, the security issues also raise concerns in the community. Adversarial attack~\cite{szegedy2013intriguing,carlini2017towards,athalye2018obfuscated,xu2019evading} is one of the most intuitive ways to evaluate the robustness of a model. In adversarial attacks, the attacker generates adversarial examples to fool a model predicting wrong. Currently, SNNs are demonstrated can be attacked through adversarial examples \cite{sharmin2019comprehensive, liang2021exploring, buchel2021adversarial,  marchisio2021dvs}. It is urgent to explore an efficient way to improve the robustness of SNN models.

Previously, researchers have investigated the impact of hyper-parameter selection \cite{el2021securing} and input filtering \cite{marchisio2021dvs} on the adversarial attack in SNNs. However, these methods do not directly promote the classification behavior of a given SNN model. Unlike SNNs, how to improve the robustness of an artificial neural network (ANN) is well studied. Recently, trianing a neural network model with certified defense methods~\cite{kolter2017provable,mirman2018differentiable,zhang2020towards,xu2020automatic} show remarkable guarantee to improve the model's robustness. CROWN-IBP \cite{zhang2020towards} is one of the most promising certified training methods with polynomial computianal cost compared with natural training. The CROWN-IBP method will compute the output boundary for a given bounded input. The core mission in CROWN-IBP certified training is to find the upper and lower bound function for each operation and find tight linear relaxation for non-linear operations. However, the current CROWN-IBP method cannot be directly applied to SNNs. Firstly, the neuron dynamic in SNNs is more complicated. Hence, some new boundary functions should be defined to bound the unique non-linear operations in SNNs. Secondly, SNNs accept both spike and digital inputs, which requires additional boundary generalization for different input types.

Enlightened by certification training in ANNs, in this work, we designed an end-to-end robust training method to improve the robustness of an SNN model against adversarial attacks. Specifically, our major contributions can be summarized as follows:
\vspace{-2mm}

\begin{itemize}[itemsep=0pt]
    \item We design S-IBP and S-CROWN to tackle the non-linear fire function and temporal update in SNNs which firstly introduce SNNs to the linear relaxation based verification family.   
    \item We formalized $\ell_0$-norm and $\ell_\infty$-norm boundaries for spike and digital inputs, respectively.
    \item Our proposed methods are evaluated on MNIST \cite{lecun1998gradient}, FMNIST \cite{xiao2017fashion} and NMNIST \cite{orchard2015converting} datasets. The experinetal results show that we can achieve a maximum $37.7\%$ attack error reduction with $3.7\%$ original accuracy loss.
\end{itemize}

\section{Preliminary and Related Work}
Considering multiple important components involved in this paper, we will provide the necessary preliminary knowledge as well as related works in this section. First, we  introduce the background of spiking neural networks (SNNs). Then we present the existing studies on the adversarial attack in SNNs. Finally, we elaborate the robust training via certification in artificial neural networks (ANNs).

\subsection{Spiking Neural Netowks (SNNs)}

\begin{figure}[ht]
\begin{center}
\centerline{\includegraphics[width=\columnwidth]{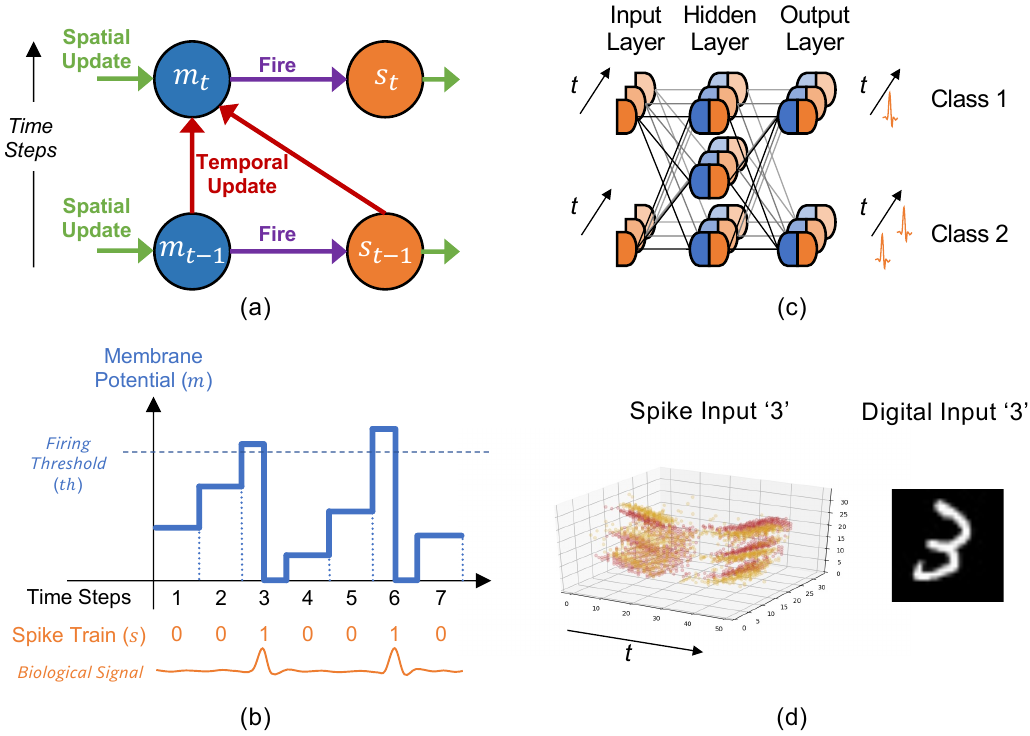}}
\caption{Introduction of SNNs: (a) neuron modeling; (b) computing model of a neuron; (c) spatiotemporal flow of an SNN model; (d) input formats of an SNN.}
\label{fig:snn_intro}
\end{center}
\vskip -0.2in
\end{figure}

\textbf{Neuron Modeling}: Usually, SNNs are designed to simulate the neuron behavior in the brain. In this work, we adopt the well-studied leaky integrated-and-fire (LIF) \cite{gerstner2014neuronal} for the neuron modeling as Figure \ref{fig:snn_intro}(a). The state of a neuron at each time step is determined by its membrane potential $m$ and its spike status $s$. Specifically, a neuron will $fire$ a spike to the post-synaptic neurons once its membrane potential is greater than a threshold $th$ and its membrane potential will be reset to 0 at the same time. An example of a neuron dynamic is shown in Figure \ref{fig:snn_intro}(b) which simulates a neuron's behavior in 7 time steps. 
The spike status of a neuron can be formulated as
\begin{equation}
    s_t[k] = fire(m_t[k]-th).
    \label{equ:spike}
\end{equation}
\begin{equation}
fire(x)=
\begin{cases}
	1, x \geq 0\\ 
	
	0, ~\text{otherwise}.
	\label{eque:fire}
\end{cases}
\end{equation}
We use $m_t[k]$ and $s_t[k]$ to represent neurons' membrane potentials and spike status at $t$-th time step and $k$-th layer. $fire(\cdot)$ is the Heaviside step function. The membrane potential of a neuron is composed of spatial and temporal update that follows
\begin{equation}
\footnotesize
    m_{t}[k]= 
    \underbrace
    {\textstyle\sum_t w[k-1] \otimes s_{t}[k-1]}_
    {spatial} +
    \underbrace
    {\alpha m_{t-1}[k] \cdot (1 - s_{t-1}[k])}_
    {temporal}.
    \label{equ:mem}
\end{equation}
The spatial update is a Convolution or Fully Connected or Pooling  (CONV/FC/POOL) operation between the weight parameters and the spike events of pre-synaptic neurons. The temporal update is determined by the neuron's membrane potential and spike status in the previous time step. Here, $\alpha$ is a scaling factor to control the decay.

\textbf{Network Structure}: Similar to ANNs, SNNs are built with input layer, hidden layers, and output layer as Figure\ref{fig:snn_intro}(c). In SNNs, the input layer receives spike inputs. The hidden layers are CONV/FC/POOL that correspond to the spatial update in the neuron modeling. Neurons in each layer propagate spike events to the neurons in the next layer. Compared to ANNs, SNNs involve an additional temporal axis that passes the neurons’ status along with time steps. The final recognition of an SNN is determined by the spike events of the output layer. In this work, we adopt the rate coding, i.e., the neuron that fires the most spikes indicates the classification result.  

\textbf{Input Format}: In this work, we focus on the image recognition tasks. The input of an SNN can be spike events captured by dynamic vision sensors \cite{patrick2008128x}, which naturally fits the input layer of an SNN. Also, SNNs can take a digital image as input, however, sampling should be involved before feeding to the SNN. In this work we adopt Bernoulli sampling \cite{deng2020rethinking} for digital inputs that follows
\begin{equation}
    P(\dot x_t[i]=1)=\hat x[i].
    \label{equ:sample}
\end{equation}
We use $\dot x_t$ to represent the spike input at time step $t$. $\hat{x}$ is the digital input after normalization to $[0,1]$ for each pixel. For a digital input, the probability that a pixel $i$ in the corresponding spike input has a spike event equals to its normalized gray value. 
The example of spike and digital inputs are shown in Figure \ref{fig:snn_intro} (d).

\textbf{Training Method}: In the early stage, researchers focus on the  SNNs trained with unsupervised local learning algorithms such as spike timing dependent plasticity (STDP) \cite{song2000competitive}. However, these learning algorithms suffer low learning accuracy and poor scalability. Recently, supervised learning via backpropagation through time  (BPTT) has been adopted in SNN training \cite{wu2018spatio}, which leverages the spatial-temporal information propagation in SNNs. Instead of achieving higher accuracy and scalability, the SNNs trained with BPTT also require fewer time steps, which is more hardware friendly in the real deployment.  Thus, in this work, we focus on the SNNs trained with BPTT-based learning method.

\subsection{Adversarial Attack in SNN}
With the development of SNNs, the security problem in SNNs attracts more attention recently. The adversarial attack is one of the most powerful threat models that affect a model's security. In this work, we will use the adversarial attack as an assessment tool to evaluate the robustness of an SNN model.

\begin{figure}[ht]
\begin{center}
\centerline{\includegraphics[width=\columnwidth]{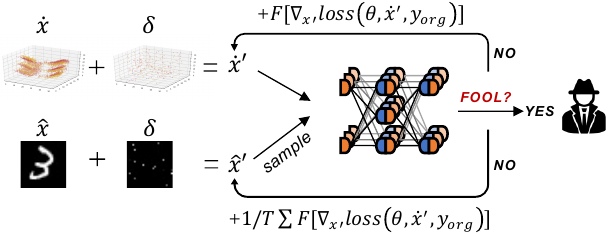}}
\caption{The framework of untargeted white-box gradient-based attack on SNN.}
\label{fig:attack_framework}
\end{center}
\vskip -0.2in
\end{figure}

\textbf{Untargeted White-box Gradient-based Attack}: 
The framework of the adversarial attack in SNNs is shown in Figure \ref{fig:attack_framework}. In adversarial attack, the attacker adds imperceptible noise on the input to fool the model generating an incorrect prediction result. In untargeted attack senario, the model predicts the adversarial example as any other class except the ground-truth, which can be formulated as
\begin{equation}
    \underset{\delta}{\text{argmin}} \lVert \delta \rVert_p,~s.t.~f(x+\delta) \neq f(x).
\end{equation}
Here, $f(\cdot)$ is the prediction result of an SNN model, $x$ is the original input, and $\delta$ is the noise added on the input. In most cases, it is easier to find an adversarial example if the attacker knows the models, i.e., the parameters and the structure of the model. The adversarial attacks that know all of the model information are called white-box attacks.

The gradient-based attacks are the most efficient ways to generate adversarial examples. In SNNs, the gradient-based attack for the spike and digital inputs can be formulated as
\begin{equation}
\begin{cases}
	\dot{x}'_{n+1} = \dot{x}'_{n} + F[\triangledown_{\dot{x}'_{n}}L(\theta, \dot{x}'_{n},y_{org})], ~\text{~~~~~~~~spike input,}\\ \\
	
	\hat{x}'_{n+1} = \hat{x}'_{n} + \cfrac{ \sum_t F[\triangledown_{\dot{x}'_{n}}L(\theta, \dot{x}'_{n},y_{org})]}{T}, ~\text{digital input}.   
\end{cases}
\end{equation}
We use $\dot{x}'$ and $\hat{x}'$ to represent the spike and digital adversarial examples, i.e., $x'=x+\delta$. $L$ and $\theta$ are the loss function and parameters of the model, respectively. The adversarial example is constructed by adding the gradient of inputs with the original label ($y_{org}$) in loss function. $F$ is a filter function that samples, clips, and generates spike compatible noise. For the digital inputs, the binary noises are averaged along the time steps ($t$) to construct the floating-point noise. During the attack, attackers can compute the adversarial examples iteratively once the current adversarial example cannot fool the model successively. We use $n$ to denote the attack iteration.


In this work, the adversarial attack works as a tool to evaluate the robustness of an SNN model. Thus, we mainly adopt the more general attack method, i.e.,untargeted white-box gradient-based attack as our assessment metric.

\subsection{Certified Training}
\label{sec:certify train}
Recently, the certified training~\cite{gowal2018effectiveness,zhang2020towards,xu2020automatic} has been demonstrated to improve the guaranteed robustness of a neural network. In this work, we leverage the CROWN-IBP method~\cite{zhang2020towards}, one of the state-of-the-art certified training that can achieve tighter bounds in accepctable training cost. Considering a digital input data bounded with $\ell_\infty$-norm, the goal of CROWN-IBP is to identify whether arbitrary input data within the boundary can fool the model. The CROWN-IBP contains two parts of bounding methods: IBP~\cite{gowal2018effectiveness} and CROWN~\cite{zhang2018efficient}.

\textbf{IBP}: 
In IBP processing, the lower and upper bounds of each layer's feature map are computed along the forward propagation, i.e. start from the input layer to the output layer. During the bound computation, the linear operation can be easily bounded once we know the maximum and minimum values of input. However, the upper and lower bound after the non-linear operations need to be dedicated to analysis. Once we acquire the bounds of output we can evaluate the robustness of the model for the given input purbation set.

\textbf{CROWN}:
Unlike IBP, CROWN bounds the model in a backward propagation manner recursively. The goal of CROWN is to formulate the output bounds as a linear equation of input. In order to achieve this goal, every operation in an NN model should be bounded by two linear equations. 

Although CROWN can achieve very tight bounds, the computinal cost is remarkably higher than IBP. So  CROWN-IBP, by combining the fast IBP bounds in a forward bounding pass and CROWN  in a backward bounding pass can efficiently and consistently outperforms IBP baselines on training verifiably robust neural networks.

\section{Methodology}
From previous studies, certification training is an efficient methodology to improve the robustness of a neural network model. However, existing methods cannot be directly applied to the SNNs. The main reason is that the non-linear function in SNNs (fire and temporal update) and the input boundary formalization are special in SNNs. In order to achieve certification training in SNNs, we designed S-IBP and S-CROWN to tackle the non-linear neuron behavior in SNNs. Also, we analyzed the input boundary for both spike and digital data.

\subsection{S-IBP \& S-CROWN}
As described in Section \ref{sec:certify train}, the core mission of certification training is to find the  bound of IBP and CROWN for each function. The information propagation in SNNs includes $fire$, temporal update and spatial update as shown in Figure~\ref{fig:snn_intro}(a). In this subsection, we detail the upper bound and lower bound of S-IBP and S-CROWN for each function. 

\begin{figure}[ht]
\begin{center}
\centerline{\includegraphics[width=\columnwidth]{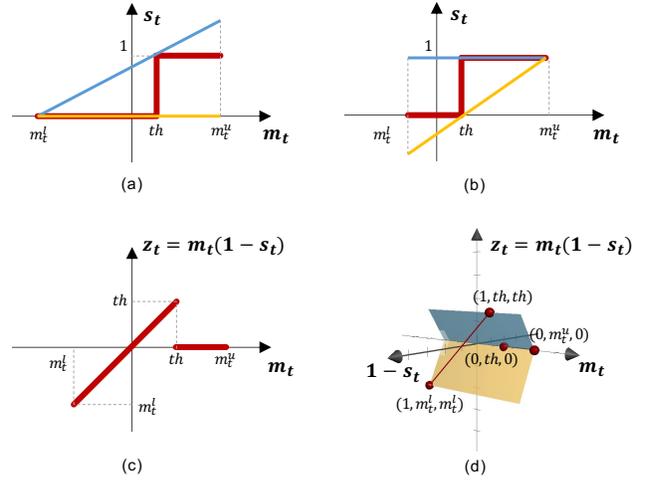}}
\caption{S-CROWN upper and lower bounds for the $fire$ function when (a) $m_t^l$ far smaller than $th$ and (b) $m_t^u$ far larger than $th$. (c) Non-linear temporal part of memory potential update. (d) S-CROWN upper and lower bounds for the temporal part of memory potential update (unstable case).}
\label{fig:fire_temporal_bound}
\end{center}
\vskip -0.2in
\end{figure}

\textbf{Fire}:
The $fire$ function describes the relation between the membrane potential $m_t$ and spike $s_t$ of a neuron as Equation \ref{equ:spike} and \ref{eque:fire}, i.e. once the neuron's membrane potential is greater than a threshold $th$, the neuron will fire a spike. Assume we have already acquired the S-IBP upper bound $m_t^u$ and lower bound $m_t^l$ for membrane potential, the S-IBP bounds for spike can be calculated with
\begin{equation}
    s_t^u = fire(m_t^u - th), ~~ s_t^l = fire(m_t^l - th).
    \label{equ:ibp_fire}
\end{equation}
During S-CROWN, we need to find two linear equations to bound the $fire$ function. When the S-IBP upper bound of membrane potential $m_t^u$ is smaller than the threshold $th$, we can conclude that the neuron must not fire. Also, when the S-IBP lower bound $m_t^l$ is greater than the threshold, we can make sure the neuron must fire.
Thus, we mainly need to consider the unstable case, i.e., $m_t^l < th 
\leq m_t^u$. In order to achieve the lowest bound relaxation, we design two boundary systems as Figure \ref{fig:fire_temporal_bound}(a) and (b). We use the red line to represent the $fire$ function. The blue and yellow lines represent the boundary functions. In our design, when the S-IBP lower bound of membrane potential $m_t^l$ far smaller than $th$, we set the S-CROWN lower bound to $s_t=0$ and the S-CROWN upper bound is a line that crosses $(m_t^l, 0)$ and $(th, 1)$. On the contrary, when $m_t^u$ is far larger than $th$, we set the S-CROWN upper bound to $s_t=1$ and the S-CROWN lower bound is a line that passes $(th,0)$ and $(m_t^u, 1)$. Overall, the S-CROWN boundary for the $fire$ function under different cases can be summarized as

\begin{equation}
    \begin{cases}
    ~0 \leq s_t \leq 0, ~~~~~~~~~~~m_t^u < th, \\\ 
    1 \leq s_t \leq 1, ~~~~~~~~~~~m_t^l \geq th, \\\ 
    
    0 \leq s_t \leq \frac{m_t - m_t^l}{th - m_t^l}, ~~ 0\leq m_t^u-th<th-m_t^l \\\
    \frac{m_t - th}{m_t^u - th} \leq s_t \leq 1, ~~~ m_t^u-th \geq th-m_t^l > 0
    
    \end{cases}
    \label{equ:crown_fire}
\end{equation}

\textbf{Temporal Update}:
According to Equation \ref{equ:mem}, the temporal update of membrane potential can be formalized as $\alpha m_{t}(1-s_{t})$. Since $\alpha$ is a constant factor, in this subsection we do not include it in the boundary analysis. Assume $z_{t}=m_t(1-s_t)$, and the S-IBP upper and lower bound for membrane potential at time step $t$ has been acquired. The relation between $m_t$ and $z_t$ can be described through Figure \ref{fig:fire_temporal_bound}(c). Based on the LIF model, the membrane potential will be reset to 0 once its value acrosses the pre-defined threshold $th$. Thus, the S-IBP bounds for $z_t$ can be computed with
\begin{equation}
    \begin{cases}
    ~z_t^u=m_t^u,  ~~ z_t^l=m_t^l, ~~~~~~~~~~~~~~m_t^u < th, \\\
    z_t^u=z_t^l=0, ~~~~~~~~~~~~~~~~~~~~~~~~~~~m_t^l \geq th, \\\
    z_t^u=th, ~ z_t^l=min(0, m_t^l) , ~~~m_t^l < th \leq m_t^u.
    \end{cases}
    \label{equ:ibp_temporal}
\end{equation}
Ituitivly, when $m_t^u < th$ or $m_t^l \geq th$, there is no boundary relaxation in S-IBP. We only need to care about the boundary when $m_t^l < th \leq m_t^u$ (unstable case). From Fgiure \ref{fig:fire_temporal_bound}(c) we can find that the $z_t\in [min(0, m_t^l), th)$ for unstable case.

During S-CROWN, $z_t$ can be also bounded without relaxation when $m_t^u < th$ and $m_t^l \geq th$. The S-CROWN upper bound (blue plane) and lower bound (yellow plane) for $z_t$ when $m_t^l < th \leq m_t^u$ is shown in Figure \ref{fig:fire_temporal_bound}(d). Here we use an additional $(1-s_t)$ axis to help us build the boundary. Note that the membrane potential is related to the spike status. Once $s_t=1 \to (1-s_t)=0$, $m_t$ must greater than $th$ and $z_t=0$. When $s_t=0 \to (1-s_t)=1$, $m_t$ must smaller than $th$ and $z_t=m_t$. Thus we need to find two planes to bound these two function (red lines in Figure \ref{fig:fire_temporal_bound}(d)). In summary, the S-CROWN boundary for the temporal update can be formulated as 
\begin{equation}
    \begin{cases}
    m_t \leq z_t \leq m_t, ~~~~~~~~~~~~~~~~~~~~~~~~~~~~~m_t^u < th, \\\
    0 ~~\leq z_t \leq 0, ~~~~~~~~~~~~~~~~~~~~~~~~~~~~~~~~m_t^l \geq th, \\\
    (1-s_t)m_t^l < z_t < (1-s_t)th, ~~~m_t^l < th \leq m_t^u.
    \end{cases}
    \label{equ:cronw_tempral}
\end{equation}
Thus, in our design, the S-CROWN boundaries for the temporal update are the functions of $s_t$ under the unstable case. 

\textbf{Spatial Update}

The spatial update in SNNs is shown in Equation \ref{equ:mem}. Similar to ANNs, the spatial update in SNNs is composed of CONV/FC/POOL (in this work we focus on the CONV and FC). In SNNs CONV/FC takes spike events and weight as input. Since the spike events are in binary format, the S-IBP of spatial update can be implemented with 
\begin{equation}
\footnotesize
    \begin{cases}
    ~center = w[k] \otimes s_t^l[k] + b[k], \\\
    sp_t^u[k+1] = center + w[k]^{+}  \otimes (s_t^l[k]=0 \cap s_t^u[k]=1), \\\
    
    sp_t^l[k+1] ~= center + w[k]^{-}  \otimes (s_t^l[k]=0 \cap s_t^u[k]=1).
    \end{cases}
    \label{equ:ibp_spatial}
\end{equation}
We use $sp_t$ to represent the result of the spatial update. $\otimes$ denotes CONV/FC operation. In Equation \ref{equ:ibp_spatial}, $s_t^l[k]=1$ represents those pre-synaptic neurons in layer $k$ who are must fire. The stable fired pre-synaptic neurons contribute the same for both S-IBP upper and lower bound of $sp_t$. The unstabled pre-synaptic neurons can be represented with ($s_t^l[k]=0 \cap s_t^u[k]=1$), whose upper and lower bound for spike status are 1 and 0. These unstable pre-synaptic neurons will perform CONV/FC with the positive and negative weights to affect the S-IBP upper and lower bound of $sp_t$. 

Since the spatial update is a linear operation, 
the S-CRWON for spatial update does not have relaxation, which is the same as the case in ANNs. During the S-CROWN phase, we do not need to design boundary functions to bound the spatial update in SNNs. 

\subsection{Input Boundary Formalization}

In SNNs, except for the distinct non-linear behavior of information propagation, the input layer of an SNN model only accept binary spike. The special data format for the input layer leads to different boundary formalizations for spike and digital images. 

\textbf{Spike Input}:
An example of spike input is shown in Figure \ref{fig:snn_intro}(d). All elements in a spike data are in binary format which is compatible with the input layer of an SNN model. For each element $\dot{x}_t[i]$ in a spike input, the boundary of that element can be either stable cases: $\dot{x}_t^u[i]=\dot{x}_t^l[i]=0$; $\dot{x}_t^u[i]=\dot{x}_t^l[i]=1$ or unstable case: $\dot{x}_t^u[i]=1, \dot{x}_t^l[i]=0$. For a spike input with uncertainty noise, we can only control how many data points in the spike input are unstable. Thus, the boundary for a spike input can be formulated with $\ell_0$-norm. Specifically, we can pick $size(\dot{x}) \times \epsilon$ elements from a spike input and set them as unstable points. Also, the $\ell_0$-norm boundary can be interpreted as the probability of an element is unstable, which can be formulated as 
\begin{equation}
\footnotesize
    P(\dot{x}_t^u[i]=1,\dot{x}_t^l[i]=0)=\epsilon ~ \Leftrightarrow ~ |\dot{x}'-\dot{x}|_0\leq size(\dot{x}) \times \epsilon
    \label{equ:spike_bound}
\end{equation}
Here, $\dot{x}'$ is an arbitrary adversarial example that has at most $size(\dot{x}) \times \epsilon$ data points different from $\dot{x}$.

Note that we can only certify the robustness of a spike input after we have picked the unstable data points. Under our robustness formulation, we cannot guarantee the robustness of a spike input under a given $\ell_0$-norm. The reason is that the search space for $\ell_0$-norm cannot be bounded. 

\begin{figure}[ht]
\begin{center}
\centerline{\includegraphics[width=\columnwidth]{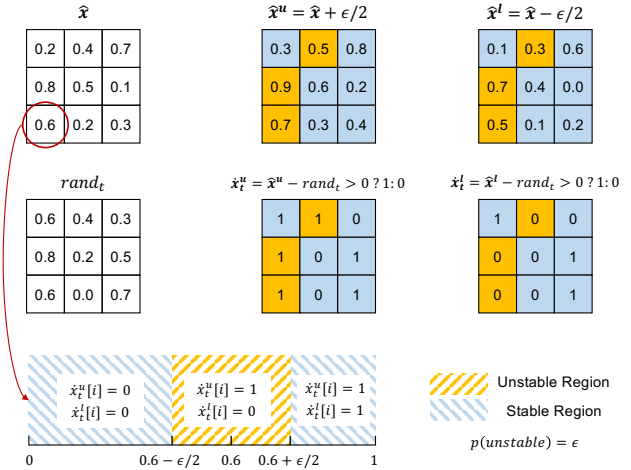}}
\caption{Relation between $\ell_\infty$-norm digital input doundary and $\ell_0$-norm spike input boundary. In this example, $\epsilon=0.2.$}
\label{fig:input_bound}
\end{center}
\vskip -0.2in
\end{figure}

\textbf{Digital Input}:

Unlike spike inputs, digital inputs need an additional Bernoulli sampling before feeding the data to the input layer as Equation \ref{equ:sample}. After the sampling, the digital input $\hat{x}$ is converted to a spike input $\dot{x}$ which fits the input layer of an SNN. Note that we have defined the $\ell_0$-norm boundary for a spike input. We need to further explore how to define a boundary for a digital input when its sampled data is bounded with $\ell_0$-norm. Based on our analysis, we find that the corresponding digital input boundary can be formalized with $\ell_\infty$-norm as Figure \ref{fig:input_bound}. We use $rand_t$ to represent a rand mask to achieve Bernoulli sampling for each time step:
\begin{equation}
    \dot{x}_t[i]=
    \begin{cases}
    1, ~~\hat{x}[i] > rand_t[i],\\
    0, ~~\text{otherwise}.
    \end{cases}
\end{equation}
After we bound the digital input with $\ell_\infty$-norm, the upper bound of digital image becomes $\hat{x}^u=\hat{x}+\epsilon/2$ and the lower bound is $\hat{x}^l=\hat{x}-\epsilon/2$. For all elements in $\hat{x}$, the difference between the upper bound and the lower bound is $\epsilon$, i.e., $\hat{x}^u[i]-\hat{x}^l[i]=\epsilon$. For each time step, $\hat{x}^u$ and $\hat{x}^l$ use the identical random map $rand_t$ to sample the corresponding spike inputs $\dot{x}_t^u$ and $\dot{x}_t^l$. Based on the sampling processing, the probability of an element in input layer is unstable can be formulated as
\begin{equation}
    P(\dot{x}_t^u[i]=1,\dot{x}_t^l[i]=0) = \hat{x}^u[i]-\hat{x}^l[i] = \epsilon.
\end{equation}
The unstable probability here is exactly the same as the case for spike input as Equation \ref{equ:spike_bound}. Thus, from the static perspective, the $\ell_\infty$-norm boundary for digital input is equivalent to the $\ell_0$-norm boundary for spike input. Since $\epsilon$ usually very small, we ignore the corner cases, i.e., when $\hat{x}[i]$ is close to 0 or 1 after normalization. 

For digital inputs, although we can bound the input with $\ell_\infty$-norm, we cannot guarantee the robustness of the input under such boundary. The reason is that during inference, the corresponding spike input is generated through sampling. Thus, for an arbitrary input, the possible sampled results are equivalent to the entire space (each element in the spike input can be 1 or 0), which cannot be bounded.

\begin{algorithm}[ht]
   \caption{S-IBP}
   \label{alg:S-IBP}
\begin{algorithmic}
\begin{scriptsize}
   \STATE {\bfseries Input:} \\
   spike input $\dot{x}$ or digital input $\hat{x}$; 
    input boundary $\epsilon$;
    robust training time steps $T'$;

   \STATE {\bfseries Func:} \\   
   
   \FOR{$t=1$ {\bfseries to} $T'$}
   
   \STATE \todor{// input boundary formalization}\\
   \IF{$\hat{x}$}
   \STATE generate random map $rand_t$;\\ 
   \STATE $\dot{x}_t^u = (\hat{x}+\epsilon/2) - rand_t ~ > ~ 0 ~ ? ~ 1:0;$\\
   \STATE $\dot{x}_t^l ~= (\hat{x}-\epsilon/2) - rand_t ~ > ~ 0 ~ ? ~ 1:0;$\\

   \ELSE
   \STATE Randomly pick $size(\dot{x}_t)\times \epsilon$ elements from $\dot{x}_t$ and label the picked elements with $pick_t$;\\
   \STATE $\dot{x}_t^u=\dot{x}_t^l=\dot{x_t}$;~~ $\dot{x}_t^u[pick_t]=1$;~~ $\dot{x}_t^l[pick_t]=0$;\\
   \ENDIF
   \STATE $s_t^u[0] = \dot{x}_t^u$; ~~$s_t^l[0] = \dot{x}_t^l$;
   
   \STATE
   \STATE inital $m_t^u[k]=m_t^l[k]=0$ for all layers.\\
   \STATE \todor{//S-IBP}
   \FOR{$k=1$ {\bfseries to} $K$} 
   \STATE \todob{//spatial update (Equation \ref{equ:ibp_spatial})}
   \STATE $m_t^u[k] += sp_t^u[k]$; ~~~$m_t^l[k] += sp_t^l[k]$
   
   \STATE \todob{//temporal update (Equation \ref{equ:ibp_temporal})}
   \IF{$t<T'$}

   \STATE $m_{t+1}^u[k] = \alpha * z_t^u$; ~~~$m_{t+1}^l[k] = \alpha * z_t^l$;
   \ENDIF
   
   \STATE \todob{//fire (Equation \ref{equ:ibp_fire})}
   \STATE $s_t^u[k]=fire(m_t^u[k]-th)$; ~~~$s_t^l[k]=fire(m_t^l[k]-th)$;
   
   \ENDFOR
   \ENDFOR


  \STATE \textbf{Return} upper and lower bound of intermediate data $\dot{x}_t^u$,  $\dot{x}_t^l$, $m_t^u$,  $m_t^l$, $s_t^u$,  $s_t^l$

\end{scriptsize}   
\end{algorithmic}

\end{algorithm}

\begin{algorithm}[ht]
   \caption{S-CROWN}
   \label{alg:s-crown}
\begin{algorithmic}
\begin{scriptsize}
   \STATE {\bfseries Input:} \\
   $\dot{x}_t^u$, $\dot{x}_t^l$, $m_t^u$,  $m_t^l$, $s_t^u$,  $s_t^l$ from S-IBP; ~~~
    robust training time steps $T'$;

   \STATE {\bfseries Func:} \\   
   
  \STATE \todor{//S-CROWN}
  \STATE build identity $I$ matrix; ~~~$As_t[K]=I/T'$;
  \FOR{$k=K$ {\bfseries to} $1$}
  \STATE initial $Am_t[k]=0$ for all time steps 
  \FOR{$t=T'$ {\bfseries to} $1$}

  \STATE \todob{//fire (Equation \ref{equ:crown_fire})}
  \STATE build $m_t[k]*d1^l+b1^l \leq s_t[k] \leq m_t[k]*d1^u+b1^u$;
  \STATE $Am_t[k] ~+=As_t[k]^- * d1^u + As_t[k]^+ * d1^l$;
  \STATE $bias ~~~~~~~+=As_t[k]^- * b1^u + As_t[k]^+ * b1^l$;
  
  \STATE \todob{//temporal update (Equation \ref{equ:cronw_tempral})}
  \IF{$t>1$}
  
  \STATE build $m_{t-1}[k]*d2^l ~+ (1-s_{t-1}[k])*d3^l ~\leq z_{t-1}[k]$;
  
  \STATE build $m_{t-1}[k]*d2^u + (1-s_{t-1}[k])*d3^u \geq z_{t-1}[k]$;
  
  \STATE $Am_{t-1}[k] = \alpha * (Am_t^-[k] * d2^u + Am_t^+[k] * d2^l)$;
  \STATE $tmp_{s} ~~~~~~~~~= \alpha * (Am_t^-[k] * d3^u + Am_t^+[k] * d3^l)$;
  \STATE $As_{t-1}[k] ~-= tmp_{s}$; ~~~~ $bias ~+= sum(tmp_{s})$;
  
  \ENDIF
  
  \STATE \todob{//spatial update};
  \STATE $As_t[k-1]=Am_t[k]*w[k-1]$; ~~~~ $bias ~+= Am_t[k]*b[k-1]$;
   
  \ENDFOR
  \ENDFOR

  \STATE \textbf{Return} $\sum_t \left( As_t[0]* (\dot{x}_t^u+\dot{x}_t^l)/2 - |As_t[0]|* (\dot{x}_t^u-\dot{x}_t^l)/2  \right ) + bias$

\end{scriptsize}   
\end{algorithmic}

\end{algorithm}

\subsection{Robust Training Algorithm}

In previous subsections, we have analyzed the S-IBP and S-CROWN for each function in SNNs. Also, we formalized the input boundaries for different input formats. Here, we present the end-to-end robust training for an SNN.

\textbf{Flexible Time Steps}: Usually, the time steps of an SNN (trained with BPTT) is 10 to 20, however, it is still large when considering robustness training. We note that for each SNN layer, the parameters (weight and bias) are shared among different time steps. Thus, we can set arbitrary time steps $T'$ for the robust training.

\textbf{S-IBP}:  During the robust training, the inputs first pass the S-IBP as Algorithm \ref{alg:S-IBP}. At the beginning, the input is bounded according to the input type. Then, the intermediate data are bounded along the forward direction. Finally, the upper and lower bound of all intermediate data are stored which will be used during the S-CROWN phase.

\textbf{S-CROWN}: Note that the goal of S-CROWN is to formulate the output bounds of a model as linear equations of input's upper and lower bounds. Also, the boundary is computed from the backward direction. The detailed steps of S-CROWN are shown in Algorithm \ref{alg:s-crown}.

In the output layer, the S-CROWN boundary can be formulated as 
\begin{equation}
    f(\dot{x}) \geq \sum_t \underbrace{I/T'}_{As_t[K]} * ~ s_t[K].
\end{equation}
Here, $I$ is the identity matrix. We use $As_t[K]$ to represent the linear matrix respect to the output, where $t$ denotes the time step and $s_t[K]$ represents the spike events in the output layer. Suppose we formulate each non-linear operation as $q_t=g(p_t)$, where $p_t$ and $q_t$ represent the input and output of non-linear function $g(\cdot)$. Based on our analysis of the $fire$ function and temporal update, we can bound each non-linear operation with linear upper and lower bounds, which can be represented as 
\begin{equation}
    p_t  * d^l + b^l \leq g(p_t) \leq p_t * d^u + b^u.
\end{equation}
Here $d$ and $b$ represent the slope and intercept of a linear function. Then, the lower bound of the output can be formulated as 
\begin{equation}
\label{equ:snn_non_linear}
    \begin{array}{l}
    f(\dot{x})  \geq  \sum_t Aq_t * g(p_t) + B \\\\
      
  ~~~~~~~~\geq \sum_t \underbrace{(Aq_t^- * d^u + Aq_t^+ * d^l)}_{Ap_t} ~ * ~p_t ~ +  \\
      ~~~~~~~~\underbrace{(Aq_t^- * b^u + Aq_t^+ * b^l + B)}_{new ~ bias ~ B}.
    \end{array}
\end{equation}
We use $A^+$ and $A^-$ to represent the positive and negative values in matrix A. Equation \ref{equ:snn_non_linear} can be used to process the $fire$ and temporal update in SNN. For the linear operation, i.e., $q_t=w*p_t+b$, the boundary propagation can be formulated as
\begin{equation}
    \begin{array}{l}
    f(\dot{x})  \geq  \sum_t Aq_t * (w*p_t+b) + B \\\\
      
  ~~~~~~~~\geq \sum_t \underbrace{(Aq_t * w)}_{Ap_t} ~ * ~p_t ~ +  \underbrace{(Aq_t * b + B)}_{new ~ bias ~ B}.
    \end{array}
\end{equation}
Now, we can follow the computation process provided in Algorithm \ref{alg:s-crown} to compute the slope matrix and bias of the bounded spike input ($\dot{x}^u$ and $\dot{x}^l$). Finally, we can follow the robust training method provided in \cite{zhang2020towards} and use the lower bound of S-CROWN to train the SNN model.


\section{Experiment}

\subsection{Experiment Setup}

\textbf{Dataset and Network Structure}: In this work, we evaluate our robust training method on three datasets: MNIST \cite{lecun1998gradient}, FMNIST \cite{xiao2017fashion} and NMNIST \cite{orchard2015converting}. MNIST and FMNIST are digital datasets and NMNIST is spike dataset. For each dataset, we use two network structures in experiments, i.e., a three-layers FC network and a four-layers CONV network. The detailed setting for datasets and network structures are shown in Table~\ref{tab:setup}. We set the firing threshold $th$ and decay factor $\alpha$ to 0.25 for the neuron modeling in Equation \ref{equ:spike} and \ref{equ:mem}.

\vspace{-4mm}
\begin{table}[ht]
\caption{Datasets and network structure}
\vskip 0.05in
\label{tab:setup}
\centering
\scalebox{0.8}{
\begin{tabular}{c|ccc}
\toprule
           & \multicolumn{1}{c|}{MNIST}   & \multicolumn{1}{c|}{FMNIST}  & NMNIST  \\ \hline
Input Type & \multicolumn{1}{c|}{digital} & \multicolumn{1}{c|}{digital} & spike   \\
Size       & \multicolumn{1}{c|}{1*28*28} & \multicolumn{1}{c|}{1*28*28} & 2*34*34 \\
Time Step  & \multicolumn{1}{c|}{10}      & \multicolumn{1}{c|}{10}      & 10      \\
FC Acc     & \multicolumn{1}{c|}{98.45\%} & \multicolumn{1}{c|}{87.58\%} & 98.30\% \\
CONV Acc   & \multicolumn{1}{c|}{99.09\%} & \multicolumn{1}{c|}{89.53\%} & 99.05\% \\ \hline \hline
FC         & \multicolumn{3}{c}{X-FC512-FC256-FC10}                                \\ \hline
CONV       & \multicolumn{3}{c}{X-C64K3S2-C128K3S2-FC256-FC10}                     \\ \bottomrule
\end{tabular}
}
\end{table}

\textbf{Original and Robust Training}: In original training, we adopt BPTT based training method \cite{wu2018spatio}. We train 80 epochs for each SNN model. During the original training, the learning rate is set to 0.01 at the beginning, it decays to 0.001 at the $55^{th}$ epoch. In robust training, we use the lower bound of S-CROWN as the loss function. During robust training, we set $\epsilon$ to 0 at the beginning.  It will increase linearly to the final $\epsilon$ during the first 250 training epochs. In the last 50 training epochs, $\epsilon$ is unchanged. 

\textbf{Adversarial Attack}: In this work we adopt the untargeted white-box gradient-based attack \cite{liang2021exploring} in SNN. In our experiment, we select 300 examples for each dataset to apply adversarial attack. In order to bound the noise of adversarial examples, we involve additional constraints during the attack. Specifically, for digital inputs, we clip the adversarial example for each attack iteration to make the adversarial example stay in the boundary. For spike inputs, each attack iteration we limit the amount of changed elements to $size(\hat{x})*\epsilon / 2$ and $size(\hat{x})*\epsilon / 6$ for FC and CONV networks to achieve the highest attack efficiency.

\subsection{Robust Training in SNNs}

\begin{figure}[htbp]
\begin{center}
\centerline{\includegraphics[width=\columnwidth]{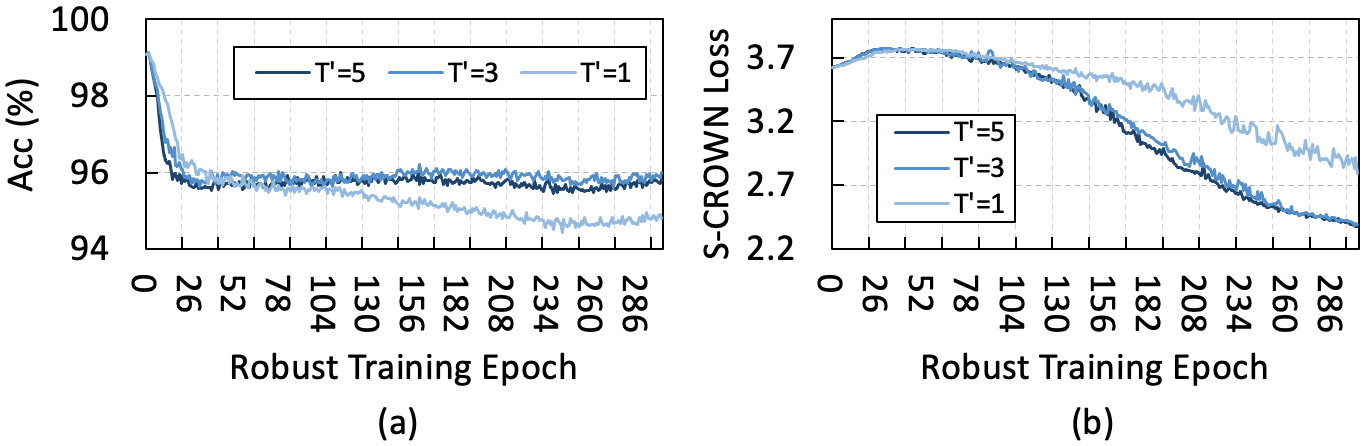}}
\vspace{-2mm}
\caption{Impact of robust training time steps $T'$ on  (a) original accuracy and (b) S-CROWN loss. (CONV model; MNIST dataset; $\epsilon=0.12$)}
\label{fig:result_t}
\end{center}
\vskip -0.2in
\end{figure}

\begin{figure*}[htbp]
\begin{center}
\centerline{\includegraphics[width=0.95\textwidth]{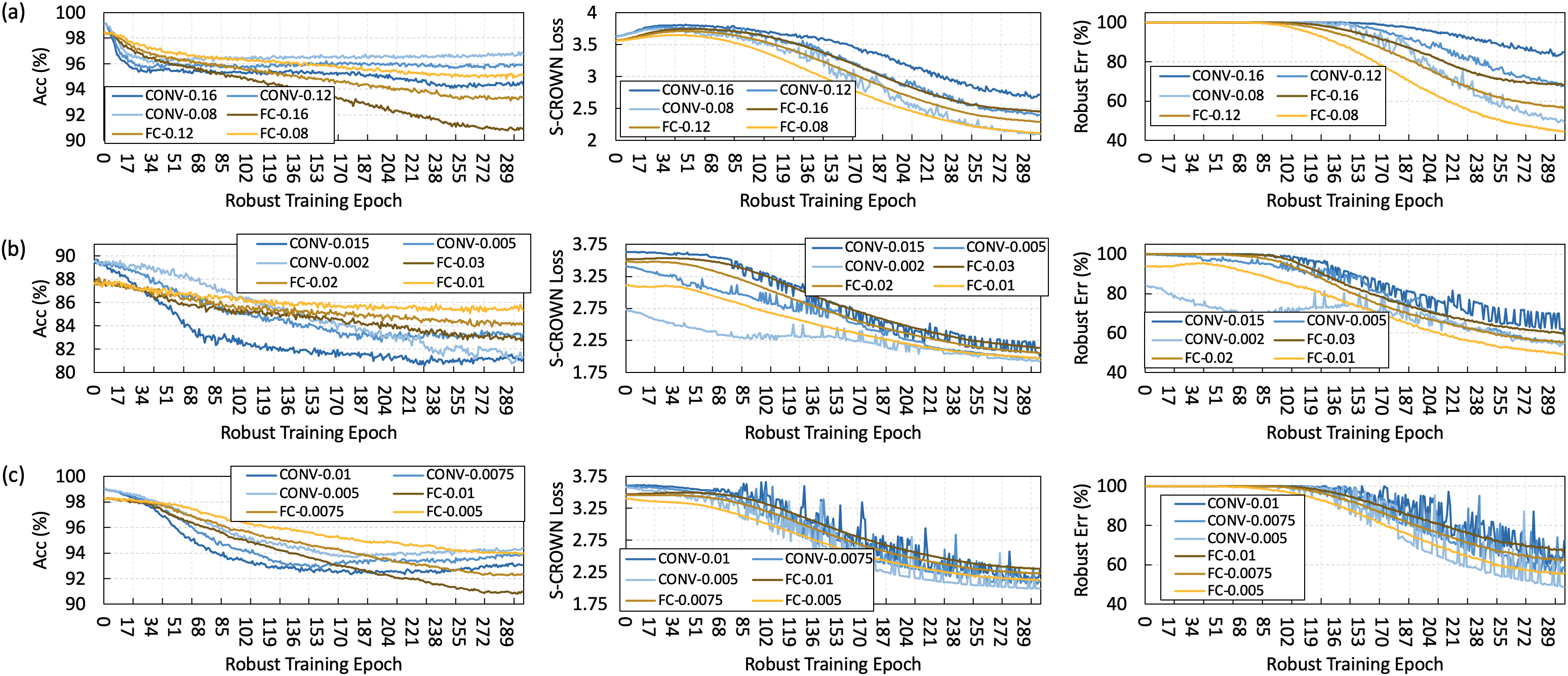}}
\caption{Original accuracy, S-CROWN loss and S-CROWN robust error during robust training under (a) MNIST; (b) FMNIST and (C) NMNIST datasets with different network structures and $\epsilon$. }
\label{fig:result_boudnary}
\end{center}
\vskip -0.3in
\end{figure*}
\vspace{-2mm}

\begin{table*}[t]

\caption{Comparing untargeted white-box gradient-based attack between the original model and the model after robustness training. }
\label{tab:attack}
\scriptsize
\scalebox{0.83}{
\begin{tabular}{c|ccccc|ccccc|ccccc}
\toprule
                      & \multicolumn{5}{c|}{MNIST (FC-$\epsilon$=0.12, CONV-$\epsilon$=0.12)}                   & \multicolumn{5}{c|}{FMNIST (FC-$\epsilon$=0.01, CONV-$\epsilon$=0.005)}                 & \multicolumn{5}{c}{NMNIST (FC-$\epsilon$=0.005, CONV-$\epsilon$=0.005)}                 \\
                      &     attack      & \multicolumn{2}{c}{original network} & \multicolumn{2}{c|}{robust network} &  attack         & \multicolumn{2}{c}{original network} & \multicolumn{2}{c|}{robust network} &  attack          & \multicolumn{2}{c}{original network} & \multicolumn{2}{c}{robust network} \\
                      & $\epsilon$       & org err          & attack err        & org err         & attack err        & $\epsilon$       & org err         & attack err         & org err         & attack err        & $\epsilon$        & org err         & attack err         & org err        & attack err        \\ \midrule
\multirow{4}{*}{FC}   & 0.104     & 0.3\%            & 19.3\%            & 4.7\%           & 16.6\% (\textcolor{cyan}{-2.7\%})           & 0.040     & 13.0\%          & 20.7\%             & 13.6\%          & 20.0\%(\textcolor{cyan}{-0.7\%})            & 0.0007     & 3.7\%           & 21.0\%             & 3.7\%          & 10.0\%(\textcolor{cyan}{-11.0\%})            \\
                      & 0.124     & 0.3\%            & 29.7\%            & 4.7\%           & 20.3\%  (\textcolor{cyan}{-9.4\%})        & 0.070     & 13.0\%          & 29.7\%             & 13.3\%          & 24.0\%(\textcolor{cyan}{-5.7\%})            & 0.0009     & 3.7\%           & 30.7\%             & 3.7\%          & 12.3\%(\textcolor{cyan}{-18.4\%})            \\
                      & 0.140     & 0.3\%            & 39.0\%            & 4.7\%           & 23.3\%(\textcolor{cyan}{-15.7\%})            & 0.100     & 11.7\%          & 41.3\%             & 14.0\%          & 32.0\%(\textcolor{cyan}{-9.3\%})            & 0.0012     & 3.7\%           & 41.3\%             & 3.7\%          & 16.7\%(\textcolor{cyan}{-24.6\%})            \\
                      & 0.154     & 0.3\%            & 50.0\%            & 5.0\%           & 24.3\%(\textcolor{cyan}{-25.7\%})            & 0.114     & 13.3\%          & 49.7\%             & 13.6\%          & 38.7\%(\textcolor{cyan}{-11.0\%})            & 0.0014     & 3.7\%           & 48.7\%             & 3.7\%          & 18.3\%(\textcolor{cyan}{-30.4\%})            \\ \midrule

\multirow{4}{*}{CONV} & 0.120     & 0.3\%            & 18.7\%            & 3.7\%           & 8.3\%(\textcolor{cyan}{-10.4\%})             & 0.040     & 10.0\%          & 19.7\%             & 13.7\%          & 22.0\%(\textcolor{cyan}{+0.3\%})            & 0.0008     & 1.3\%           & 20.0\%             & 4.3\%          & 7.0\%(\textcolor{cyan}{-13.0\%})             \\
                      & 0.140     & 0.3\%            & 29.0\%            & 4.0\%           & 9.0\%(\textcolor{cyan}{-20.0\%})             & 0.060     & 10.3\%          & 28.3\%             & 17.3\%          & 23.0\%(\textcolor{cyan}{-5.3\%})            & 0.0010     & 1.3\%           & 33.7\%             & 4.3\%          & 11.0\%(\textcolor{cyan}{-22.7\%})            \\
                      & 0.170     & 0.3\%            & 40.7\%            & 3.3\%           & 10.7\%(\textcolor{cyan}{-30.0\%})            & 0.080     & 10.0\%          & 41.3\%             & 17.0\%          & 25.7\%(\textcolor{cyan}{-15.6\%})            & 0.0012     & 1.3\%           & 41.0\%             & 4.3\%          & 11.0\%(\textcolor{cyan}{-30.0\%})            \\
                      & 0.190     & 0.3\%            & 50.7\%            & 4.0\%           & 13.0\%(\textcolor{cyan}{-37.7\%})            & 0.100     & 10.3\%          & 52.3\%             & 16.7\%          & 34.3\%(\textcolor{cyan}{-18.0\%})            & 0.0015     & 1.3\%           & 49.0\%             & 4.3\%          & 12.0\%(\textcolor{cyan}{-37.0\%})            \\ \bottomrule
\end{tabular}
}
\vskip -0.2in
\end{table*}

\textbf{Selection of Robust Training Time Steps $T'$}:  We do not follow the original time steps during the robust training. Because the spatial computations in SNNs share the same parameters between different time steps, we can use arbitrary robust training time steps $T'$. In Figure \ref{fig:result_t} we analyzed the impact of $T'$ on original accuracy and S-CROWN loss. Note that we would like to keep a higher original accuracy but reduce the S-CROWN loss after the robust training. From the result, we can find that $T'=3$ gives the optimal solution. It implies that a too smaller $T'$ cannot capture the temporal dynamics of SNNs and a larger $T'$ may cause the boundary functions in S-CROWN to become too loose. Also we find that the robust training time for each epoch is $2.6\times$, $7.7\times$, and $12.9\times$ with respect to the original plain training when we set $T'$ to 1, 3, and 5. Thus, the selection of $T'$ also influences the robust training efficiency. In the rest of our evaluations, we set $T'=3$ for all robust training.

\textbf{Robust Training on Variaous Datasets}: The analysis of robust training on different datasets with various $\epsilon$ and network structures is shown in Figure \ref{fig:result_boudnary}. From the result, we have the following observations: 1. During the robust training, it is more stable for a fully connected network(yellow curves are smoother than blue curves) because of the simpler network structure. 2. With a larger $\epsilon$, the original accuracy is dropped more after the robust training. Also, it is more difficult to achieve a smaller robust error with larger $\epsilon$. 3. In the MNIST dataset, when the network structure is CONV, the robust training is more stable as $\epsilon$ becomes larger. The potential reason is that when $\epsilon$ is small, the flipped regions in spike inputs are more diverse. 4. In FMNIST and NMNIST datasets, when the network structure is CONV, the robust training is much more fluctuating. For FMNIST dataset, the unstable may be caused by the more complicated input data (cloths) and the lower convergence in original training (final accuracy is 89.53\%). For NMNIST dataset, the unstable may come from the larger input data size. A larger input size indicates the potential input regions that can be attacked becomes more. By considering the original accuracy and S-CROWN result after the robust training, we select the $\epsilon$ as shown in Table \ref{tab:attack}. We use FC-$\epsilon$ and CONV-$\epsilon$ to represent the noise boundary we selected for different network structures during robust training.

\subsection{Network Robustness Evaluation}

We use untargeted white-box adversarial attack to compare the robustness between the original model and the model after robust training. The robustness comparison is shown in Table \ref{tab:attack}. We use the original error rate to reflect the accuracy of a model on the 300 test data. The original error may have variance during the evaluation for the digital image dataset because of the input sampling mechanism. In our experiment, we select the attack $\epsilon$ to achieve approximately 20\%, 30\%, 40\%, and 50\% attack error rate on the original model. From the result, we can find that after the robust training, the models are harder to be attacked with adversarial example in all cases. We also find that the model after robust training is more secure when the attack $\epsilon$ is larger, even though the $\epsilon$ for robust training is far smaller. The potential reason is that the binary behavior of the spike events causes the boundary propagation to diverge quickly, which makes the final boundary cover more noisy inputs. Finally, we find that model robustness can be improved more when the network structure is CONV, since more paramters can be adjusted under the CONV model. From the result, we find that the CONV model in MNIST achieves the highest robustness improvement, i.e. the attack error rate reduces $37.7\%$ with $3.7\%$ original accuracy loss when the attack $\epsilon$ equals 0.190.

\section{Discussion \& Conclusion}
For ANNs, training a neural network with certified methods not only shows remarkable efficiency to improve the model's robustness but also presents flexibility to cooperate with other robust training methods. In this work, we aim to design an efficient robust training method for SNNs based on the certified methods. Specifically, we design S-IBP and S-CROWN to tackle the distinct non-linear neuron behaviors in SNNs. Also, we formulate the input boundary for different input types. We evaluate the models' robustness to untargeted white-box adversarial attack. Based on our results, we can achieve at most $37.7\%$ attack error reduction with $3.7\%$ original accuracy loss, which demonstrates the efficiency of our proposed method.


\bibliography{example_paper}
\bibliographystyle{icml2022}



\end{document}